\documentclass[runningheads]{llncs}
\usepackage[T1]{fontenc}
\usepackage{graphicx}
\usepackage{caption}
\usepackage{subcaption}
\usepackage[colorlinks]{hyperref}
\usepackage[svgnames]{xcolor}
\definecolor{highlightgrey}{rgb}{0.93,0.93,0.93} 
\NewDocumentCommand{\codeword}{v}{\colorbox{highlightgrey}{\texttt{#1}}}

\begin{document}
%
\title{PyEvalAI: AI-assisted evaluation of Jupyter Notebooks for immediate personalized feedback}

\titlerunning{PyEvalAI: AI-assisted evaluation of Jupyter Notebooks}
%

\author{
Nils Wandel
\and
David Stotko
\and
Alexander Schier
\and
Reinhard Klein
}
\authorrunning{N. Wandel et al.}
%
\institute{
Rheinische Friedrich-Wilhelms-Universität Bonn, Bonn NRW, Germany \\
\email{wandeln@cs.uni-bonn.de}
}
%
\maketitle              
\begin{abstract}




Grading student assignments in STEM courses is a laborious and repetitive task for tutors, often requiring a week to assess an entire class. 
For students, this delay of feedback prevents iterating on incorrect solutions,  hampers learning, and increases stress when exercise scores determine admission to the final exam. 

Recent advances in AI-assisted education, such as automated grading and tutoring systems, aim to address these challenges by providing immediate feedback and reducing grading workload. 
However, existing solutions often fall short due to privacy concerns, reliance on proprietary closed-source models, lack of support for combining Markdown, LaTeX and Python code, or excluding course tutors from the grading process. 


To overcome these limitations, we introduce \textbf{PyEvalAI}, an AI-assisted evaluation system, which automatically scores Jupyter notebooks using a combination of unit tests and a locally hosted language model to preserve privacy. 
Our approach is free, open-source, and ensures tutors maintain full control over the grading process. 
A case study demonstrates its effectiveness in improving feedback speed and grading efficiency for exercises in a university-level course on numerics. 

\end{abstract}

\section{Introduction}

Weekly student assignments play an important role in STEM (Science Technology Engineering Mathematics) education as they help students to apply curriculum content in practice. 
Furthermore, exercise sheets motivate students to learn on a regular basis 
and are a good predictor for the final exam performance \cite{El-Hashash_2022,kim2022homework,latif2020impact}. 
%
A widely used option to create exercise sheets are Jupyter notebooks \cite{BASCUNANA2023155,9648674,9782924}, since they provide an easy way to combine Markdown and Latex instructions with interactive Python code. 
However, grading and providing feedback on student assignments is a cumbersome and repetitive task for teaching assistants, typically taking around a week to complete for an entire class. 
For students, such a delay of feedback prevents iterating on incorrect solutions and hampers the learning experience. 

On the other hand, recent breakthroughs in large language models are currently transforming traditional education in groundbreaking ways. 
LLM-based solutions like ChatGPT already pass the bar exam for lawyers \cite{katz2024gpt}, state examinations for medicine in Germany and the UK \cite{singhal2025toward,lai2023evaluating} or exams in micro- and macro-economics \cite{geerling2023chatgpt}.
In STEM subjects such as computer science \cite{bordt2023chatgpt}, physics \cite{pimbblet2024can,revalde2025can} or warehousing studies \cite{computers14020052}, LLMs are often en par or exceed human baselines. 
Lately, OpenAI's o3 models outperformed most humans in coding and maths competitions \cite{openaio3mini}. 
Given these impressive general problem solving capabilities, it is a logical next step to apply LLMs also in the context of education \cite{wang2024largelanguagemodelseducation} and help students to solve problems on their own. 
The immediate feedback of LLMs has not only the potential to provide a more interactive learning experience for students \cite{khanmigo} but can also relieve tutors of some of the repetitive grading work \cite{personify}. 
Unfortunately, there are still several points that must be considered when incorporating AI systems in education such as data privacy concerns, reliance on expensive proprietary closed-source software, grading accuracy and consistency, hallucinations, and adversarial attacks \cite{muehlhoff2025chatbotsimschulunterrichtwir}.

Therefore we introduce \textbf{PyEvalAI}, a novel and open-source\footnote{All code and prompts for this project will be released as open source under terms of the MIT licence upon acceptance. } AI-assisted evaluation system that automatically generates scores and feedback for Jupyter assignments using a combination of unit tests and a language model hosted by the university itself.
While the AI generated feedback provides a more interactive learning experience for students and reduces the workload for tutors, tutors can still easily correct inaccurate or inconsistent AI feedback at any time to mitigate the current shortcomings of LLMs. 

This paper is structured as follows: First, we cover related work. Next, we explain the architecture of PyEvalAI as well as its user interface and back-end in more detail. Then, we present the results of a case study in context of a university-level numerics course and conclude our work.

\section{Related Work}

\subsubsection{Commercial Tools}
of numerous educational start-ups nowadays incorporate AI-based features in their products. For example, Khanmigo by Khan academy \cite{khanmigo} offers students interactive learning experiences by chatting with historical figures or by giving hints on questions without giving away the answer. Personify \cite{personify} allows teachers to create tailored teaching assistants for their courses. myTAI \cite{myTAI} helps teachers create teaching materials, and tools like gotFeedback \cite{gotFeedback}, Fellofish \cite{Fellofish} or Fobizz \cite{fobizz} help with automatic grading and feedback for writing exercises. Cocalc \cite{cocalc} and Vocareum \cite{vocareum} integrate generative AI into collaborative notebooks to help students. 
However, commercial tools usually outsource the AI to external providers such as OpenAI potentially leading to conflicts with privacy policies of universities. 
Furthermore, licenses can be quite expensive for lower-budget universities and students. 
Finally, it is important to keep tutors in charge of the final grades since the quality of AI-only feedback without human oversight is often still insufficient \cite{muehlhoff2025chatbotsimschulunterrichtwir}.

\subsubsection{Open-source tools} 
have been developed on multiple occasions to facilitate grading Jupyter Notebooks. Otter-Grader \cite{ottergrader} as well as OKpy \cite{okpy} were implemented at UC Berkley to serve in computer science and data science courses. UNCode \cite{su132112050} provides immediate feedback for students of an introductory lecture for "Intelligent Systems and Machine Learning" and NB grader \cite{Jupyter2019} is widely used for creating and grading assignments in computer science. 
However, these tools primarily focus on testing code solutions with rigorous unit-tests and do not support grading text exercises (for example mathematical derivations). Grading such exercises requires a more flexible LLM as a backbone.

\subsubsection{Large language models} like ChatGPT (OpenAI), Gemini (Google), Claude (Anthropic) have shown tremendous success in recent years and offer convenient APIs for easy access in the cloud. However, sending sensitive student data to external providers without explicit consent of the students might violate privacy policies of the university. 
Fortunately, more and more powerful open weight models become publicly available, such as Llama \cite{touvron2023llama_1,touvron2023llama_2,dubey2024llama3herdmodels}, Mistral \cite{jiang2023mistral7b,mistralnemoblog,mistrallarge}, 
DeepSeek \cite{deepseekai2025}, and many more. Such models can run on local infrastructure to avoid any privacy concerns.





\section{PyEvalAI}
\begin{figure}[h]
    \centering
    \includegraphics[width=0.8\textwidth]{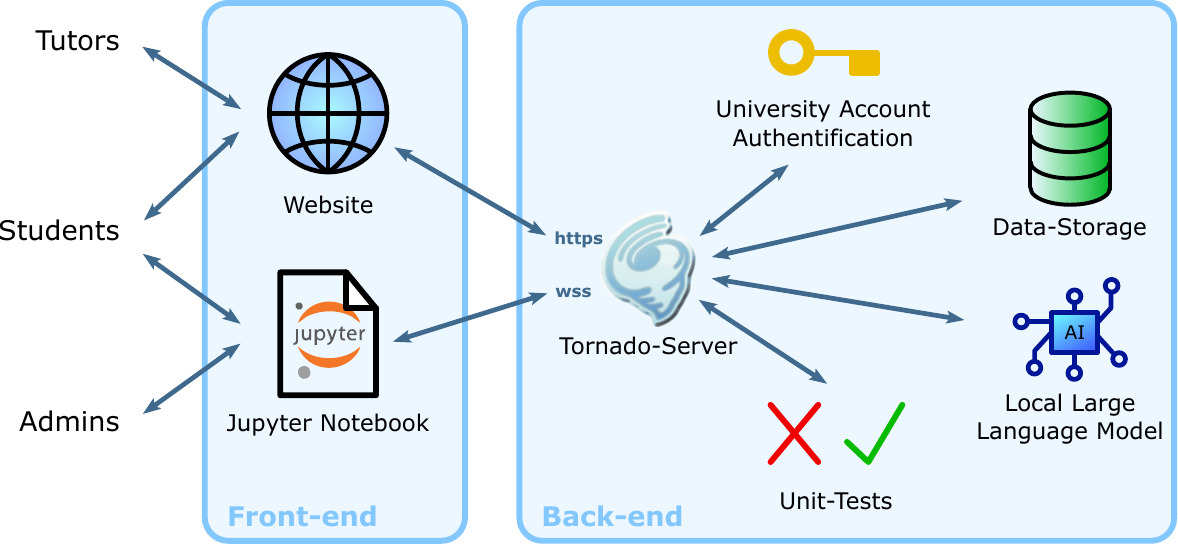}
    \caption{Architecture of PyEvalAI. Section \ref{sec:user_interface} describes the front-end, Section \ref{sec:backend} provides details about the back-end.} 
    \label{fig:architecture}
\end{figure}

The core of PyEvalAI consists of a Tornado server. As shown in Figure \ref{fig:architecture}, this server provides a front-end for students, tutors and administrators (more information is given in Section \ref{sec:user_interface}) and coordinates different back-end services such as user authentification, data-storage, unit-tests or the local LLM to generate automated feedback (see Section \ref{sec:backend}).

\subsection{User-Interface / Front-end}
\label{sec:user_interface}

The user interface of PyEvalAI allows for two different kinds of interactions: First, a web-interface for tutors and students to review grades,
and second, Jupyter Notebooks for students and administrators to work on, hand in and register exercises on the server via WebSockets \cite{rfc6455}. In the following sections, we explain the different interfaces for students, tutors and administrators in detail.

\subsubsection{Student Interface:}
\label{sec:student_interface}

\begin{figure}
\centering
\begin{minipage}{.48\textwidth}
  \centering
  \includegraphics[width=1\linewidth]{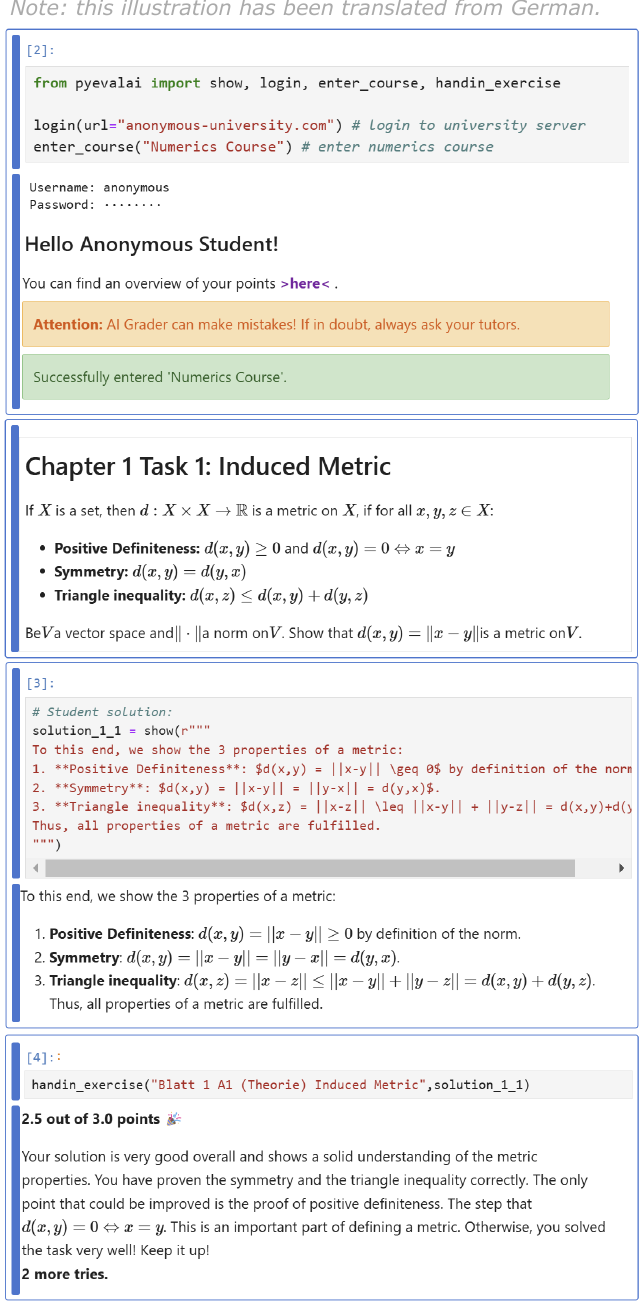}
  \captionof{figure}{Example of Jupyter notebook for students. 1. block: login and enter numerics course. 2. block: exercise description. 3. block: student solution. 4. block: hand in exercise and obtain feedback by pyevalai. For best clarity, please view this figure digitally and zoom in as needed.}
  \label{fig:jupyter_students}
\end{minipage}%
\hspace{0.03\textwidth}
\begin{minipage}{.48\textwidth}
  \centering
  \includegraphics[width=1\linewidth]{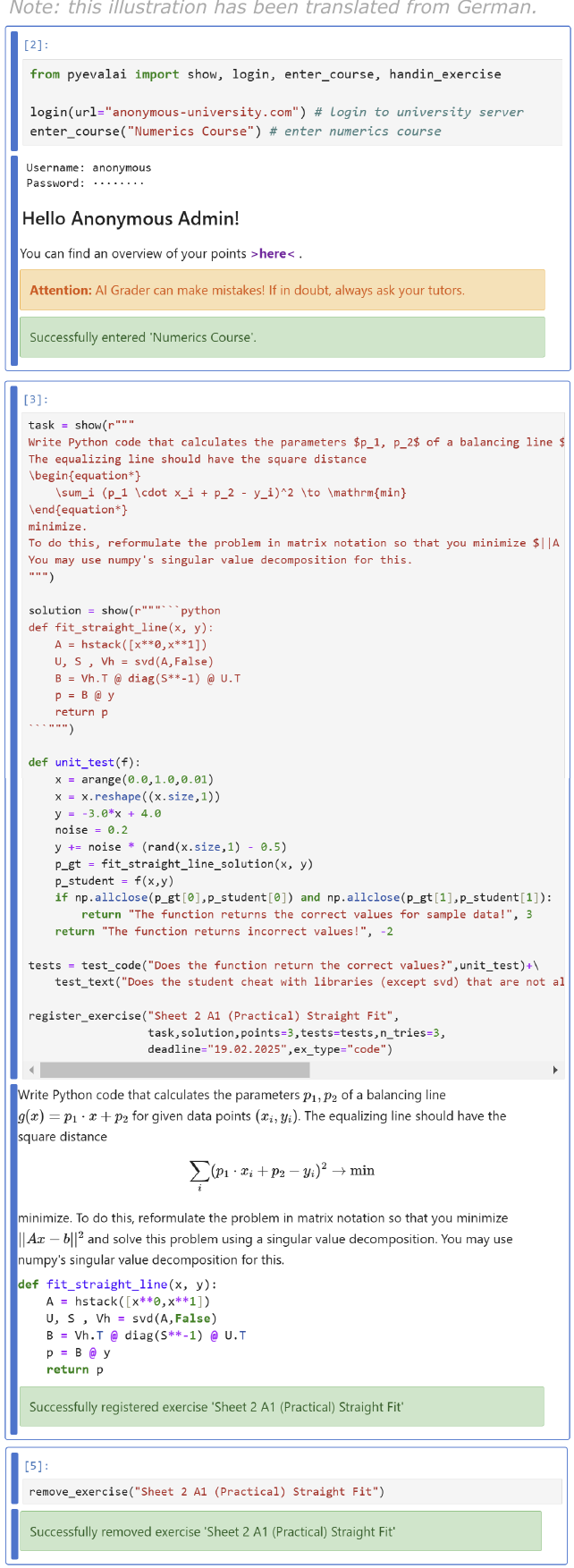}
  \captionof{figure}{Jupyter notebook for admins. 1. block: login and enter course. 2. block: specify task, solution, unit tests and register exercise. 3. block: remove exercise}
  \label{fig:jupyter_admins}
\end{minipage}
\end{figure}

Assignment sheets are given to students in the familiar form of Jupyter notebooks. Figure \ref{fig:jupyter_students} shows how students can access all the necessary functionality to hand in exercises through the pip package \codeword{pyevalai}.
%
This module provides a simple API with functions like \codeword{login("server.url")} for signing in, entering a course via \codeword{enter_course("course name")} and handing in exercises via \codeword{handin_exercise("exercise name", solution)}. 
Depending on the exercise type, students can hand in text solutions given as a string 
or code solutions given as a python function. 
%
After handing in an exercise, PyEvalAI needs 
about 1-2 minutes (depending on the exercise type and current load of the server) to generate feedback and compute a grade which is directly displayed inside the Jupyter Notebook. 
The \codeword{handin_exercise} call is asynchronous, allowing students to proceed working on subsequent exercises without having to wait for the feedback of prior exercises to be computed.


\begin{figure}[h!]
\centering
\begin{minipage}{.48\textwidth}
  \centering
  \includegraphics[trim={0 18cm 0 0},width=1\linewidth,clip]{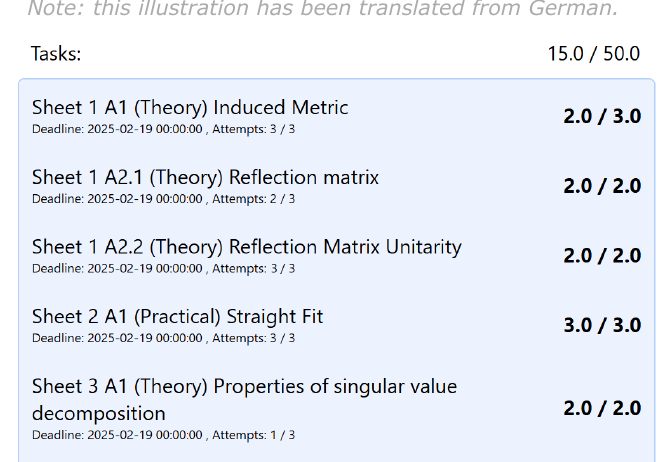}
  \captionof{figure}{Students can easily overview all exercises, already achieved scores, corresponding deadlines and numbers of attempts.}
  \label{fig:student_overview}

  \includegraphics[trim={0 2cm 0 0},width=1\linewidth,clip]{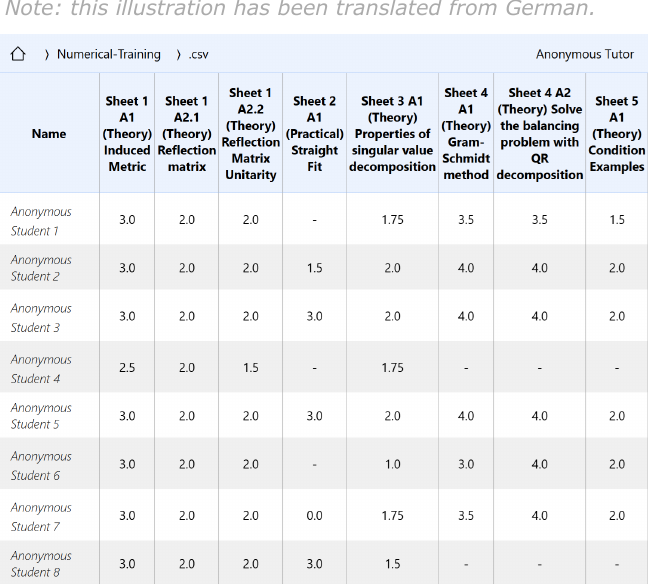}
  \captionof{figure}{Tutors can oversee in real-time all grades achieved by the students for the individual assignments in a table. By clicking on a grade, tutors can access further details and fix incorrect grades (see Figure \ref{fig:exercise_tutor}).}
  \label{fig:table_tutor}
  
\end{minipage}%
\hspace{0.03\textwidth}
\begin{minipage}{.48\textwidth}
  \centering
  \includegraphics[width=1\linewidth]{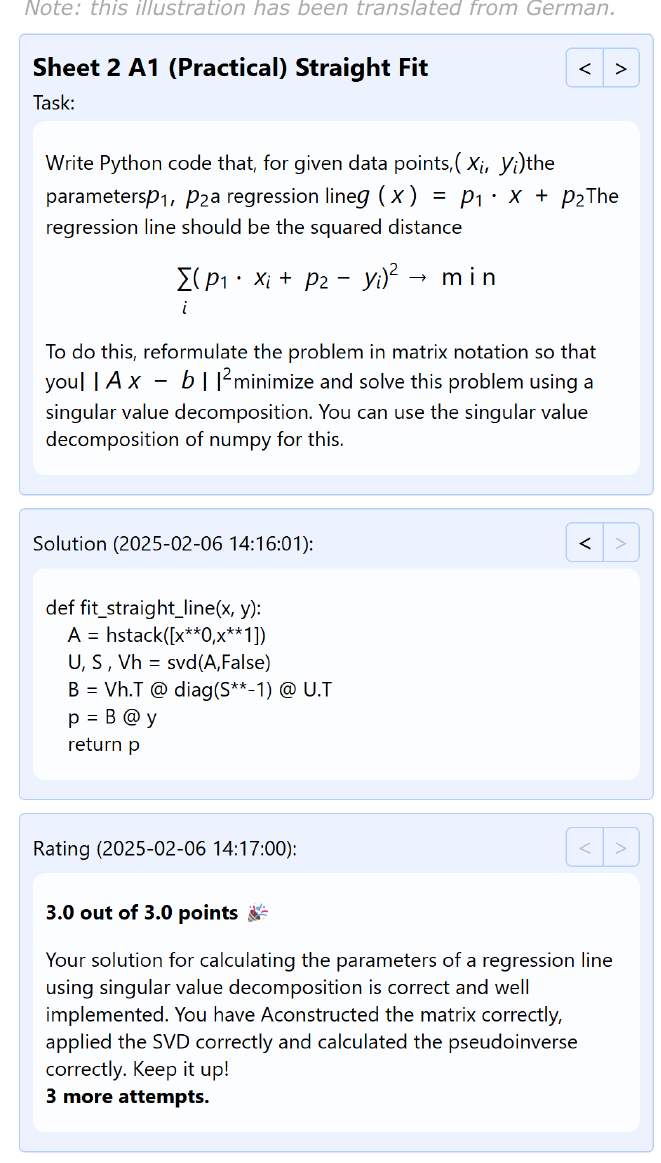}
  \captionof{figure}{For every exercise, students can transparently examine their handed in solutions as well as grades provided by PyEvalAI or tutors.}
  \label{fig:exercise_student}
\end{minipage}
\end{figure}

To get an overview of all exercises from all exercise sheets together with the achieved grades, numbers of attempts and deadlines in one place, students can visit the PyEvalAI website (see Figure \ref{fig:student_overview}). By selecting a specific exercise, students can recap the task assignment, all handed-in solution attempts and view corresponding feedback and grades provided by the LLM and tutors (see Figure \ref{fig:exercise_student}).


\subsubsection{Tutor Interface:}
\label{sec:tutor_interface}

\begin{figure}[htp!]
\centering
\begin{minipage}{.48\textwidth}
  \centering
  \includegraphics[width=1\linewidth]{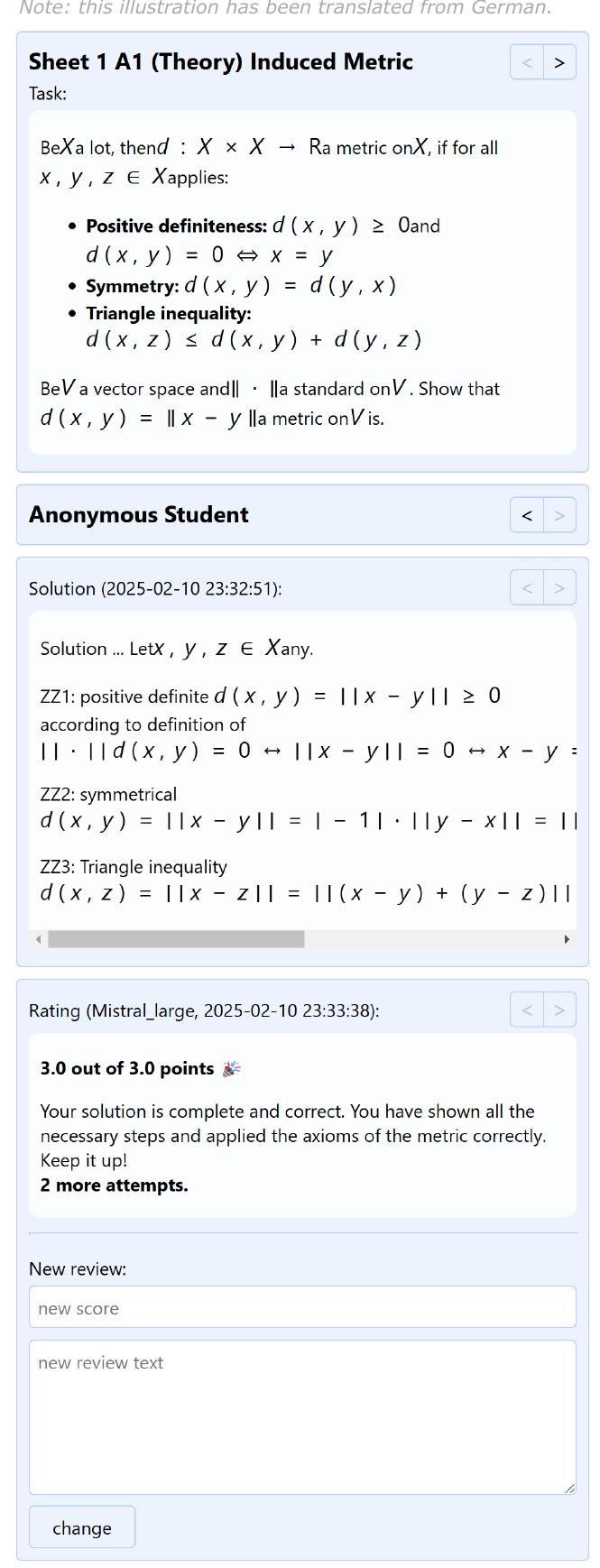}
  \caption{Tutors can review all submitted assignments and feedback generated by PyEvalAI. Feedback that does not meet the required quality standards can be easily overwritten by the tutors.}
  \label{fig:exercise_tutor}
\end{minipage}
\hspace{0.03\textwidth}
\begin{minipage}{.47\textwidth}
  \centering
\begin{subfigure}[t]{1\textwidth}
    \includegraphics[width=1\linewidth]{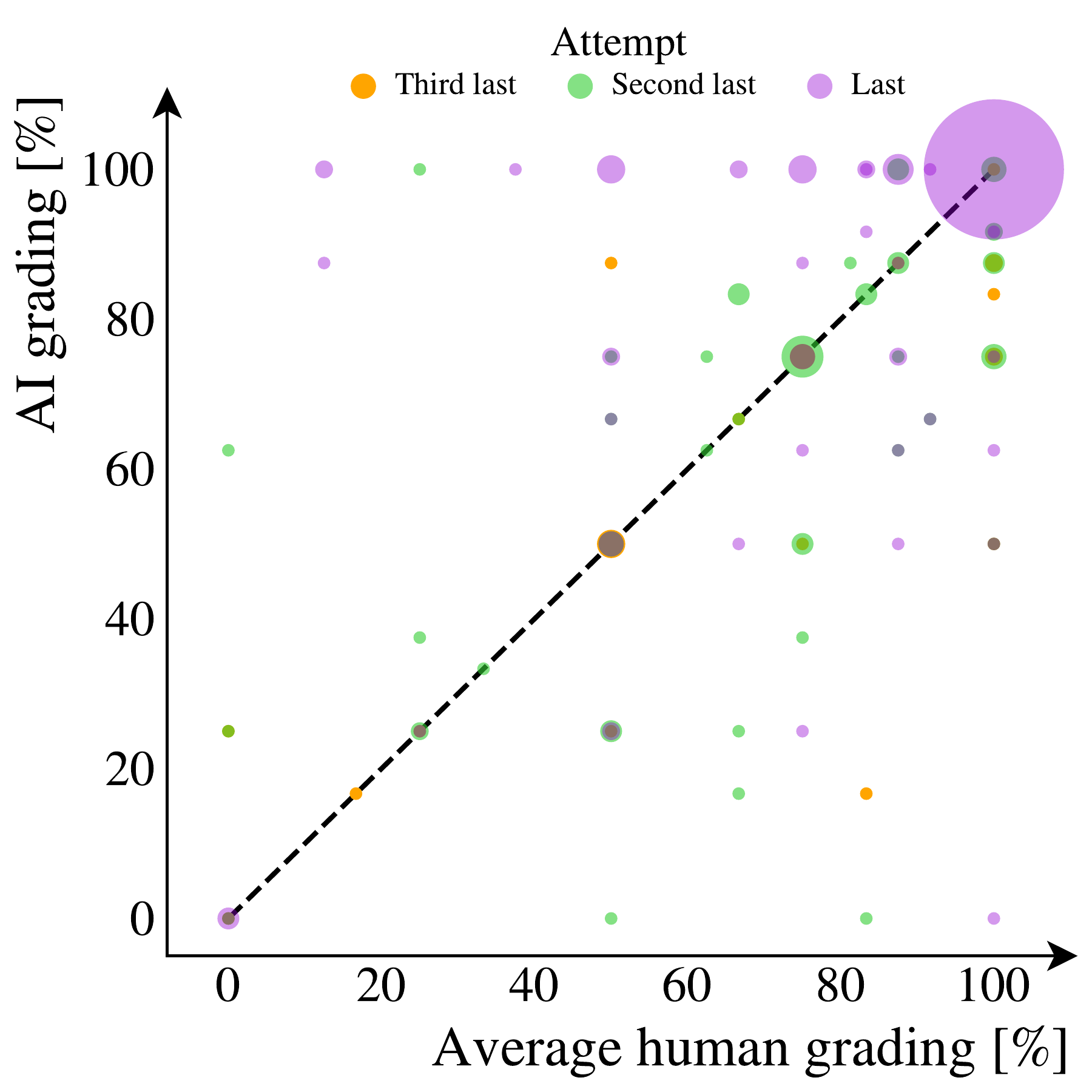}
    \caption{This scatter plot shows a high correlation of AI grades and human grades.
    The point sizes indicate the number of occurences. }
    \label{fig:grading_comparison}
\end{subfigure}
\begin{subfigure}[t]{1\textwidth}
    \includegraphics[width=1\linewidth]{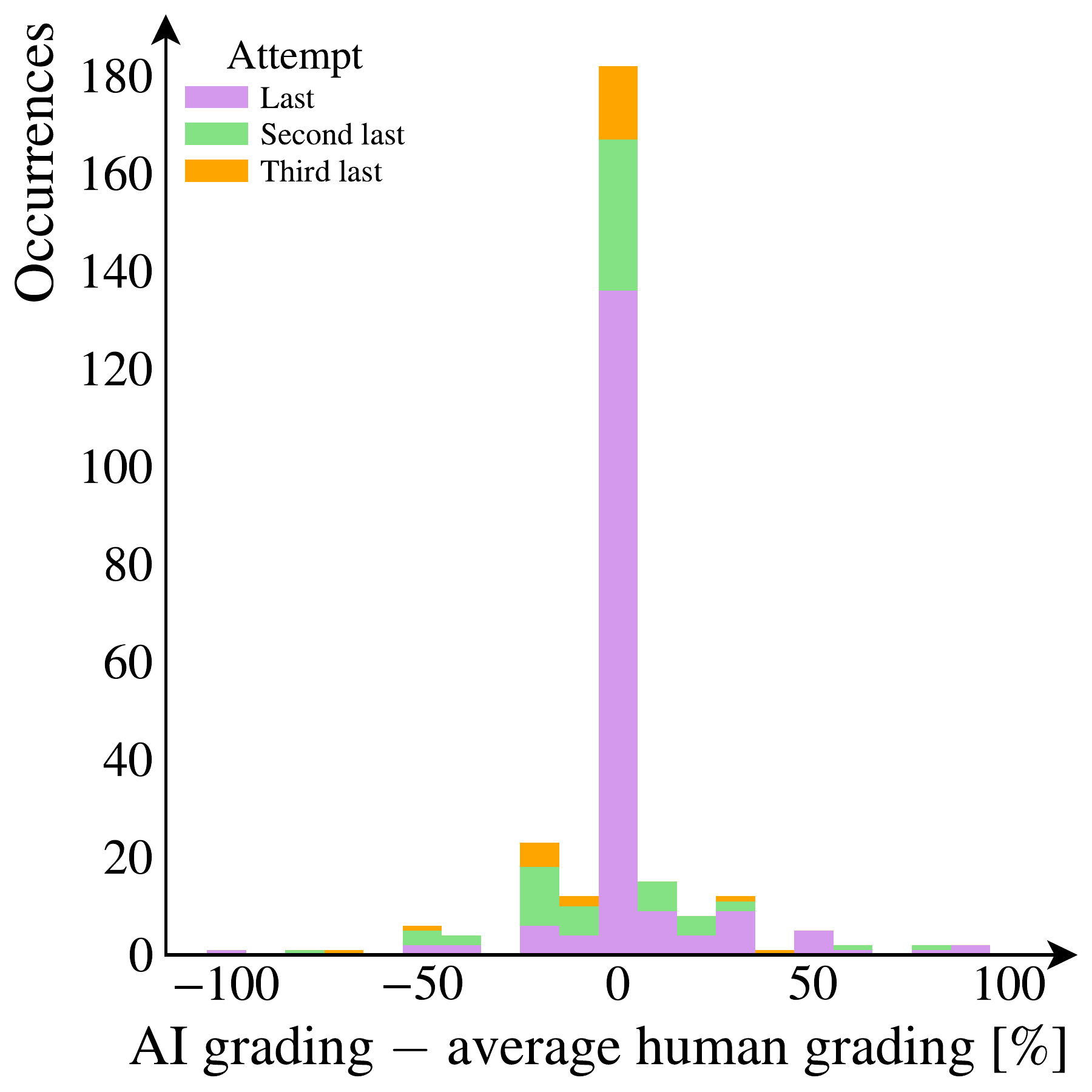}
    \caption{This histogram shows how differences between AI and human grades are distributed. Most solutions were graded the same by the AI and human tutors.
    The mean difference is $-0.14\%$ with a standard deviation of $20.73\%$.}
    \label{fig:grading_differences}
\end{subfigure}
\caption{Comparison of AI grades with average grades provided by human tutors.}
\label{fig:comparison_plots}
\end{minipage}%
\end{figure}

PyEvalAI provides tutors with a comprehensive table that contains grades for all students and all exercises (see Figure \ref{fig:table_tutor}). This allows to immediately identify exercises with poor grades that need special attention in subsequent tutorials. 
By selecting a table entry, tutors can view the task assignment, the handed-in solution attempts by the student as well as the corresponding achieved grades and feedback (see Figure \ref{fig:exercise_tutor}). If the AI feedback is inadequate, tutors can easily rectify incorrect grades and adjust the feedback.





\subsubsection{Administrator Interface:}
Course administrators have special privileges to register new exercises or remove existing exercises from a course. 
Typically, this is the role of
a teaching assisstant
or the professor who supervises the tutors. 
The interface for registering new exercises or removing existing ones is similar to the student interface for handing in exercises. 
As shown in Figure \ref{fig:jupyter_admins}, administrators create a Jupyter Notebook, import the \codeword{pyevalai} module, login with their credentials and enter the course they want to work on. 
Administrators can then register new exercises or update existing exercises with \codeword{register_exercise()}. Here, they can specify parameters like the name of the exercise, task description, sample solution, maximum number of achievable points, number of attempts (in our case, students always had 3 attempts), deadline or whether the exercise is a "text" exercise expecting a string for the solution or a "code" exercise expecting a python function. On top of that, a list of (unit-) tests can be provided for a more detailed specification of the evaluation criteria (more information on that is provided in Section \ref{sec:backend}). 
By calling \codeword{remove_exercise("exercise name")}, administrators can remove existing exercises.


\subsection{Assignment grading / Back-end}
\label{sec:backend}

The back-end of PyEvalAI consists of multiple components that are coordinated by a Tornado server. 
Tornado is a lightweight and open source python web framework that can handle many concurrent connections efficiently. For user authentication, the "Lightweight Directory Access Protocol" (LDAP) is used to check login credentials of students, tutors and administrators affilitated with the university. The datastorage consists of a Pickle file which is updated whenever new data comes in and loaded whenever the server restarts. This simple approach yields sufficient performance and facilitates handling the data in python for later evaluation and dataset generation tasks. To grade the handed in solutions, unit tests and a large language model come into play, which we'll discuss in the following sections in more detail.


\subsubsection{Large Language Model (LLM)}
To ensure data privacy for the students, the language model is hosted locally on a single NVidia A100 GPU on an in-house GPU server. 
For inference, we rely on the "Text Generation Inference" (TGI) toolkit by Huggingface \cite{huggingface_tgi} as it provides fast and efficient text generation and provides the commonly used OpenAI API for accessing text generation. 
This gives us full flexibility for different models that may come up in the future. 
During the initial development phase, we tested several models such as quantized versions of LLaMa \cite{dubey2024llama3herdmodels} and distilled versions of Deepseek \cite{deepseekai2025} 
but found that an AWQ-quantized version \cite{lin2024awq} of Mistral Large \cite{mistrallarge} produced qualitatively the most convincing results while running at acceptable inference speed.
Using Mistral Large, PyEvalAI took 88.2 seconds on average to compute feedback for a submission and around half of all responses were computed within 1 minute. 

To grade exercises and generate feedback, we first provide the model with the task description, a sample solution and the student solution. 
The LLM is then prompted with optional (unit-) test questions, which are described in more detail in the next section, to check for correctness and award partial points. 
Then, we ask the LLM to carefully compare the sample and student solution and comment on the severeness of detected differences. 
Furthermore, we ask the LLM if critical steps from the sample solution were ommited and, if so, request details on what was missed.
Finally, the LLM is tasked with determining a final score based on these insights and generating a concise, encouraging feedback message that explains the points awarded and any deductions. 
To achieve deterministic outputs, we use greedy sampling. 

This step-by-step approach, combined with a detailed comparison between the sample and student solutions, generates precise and constructive feedback for the student and ensures that feedback is both comprehensive and actionable, giving students clear insights into their work and concrete steps for improvement.

\subsubsection{Unit-Tests}
By specifying unit tests (see Figure \ref{fig:jupyter_admins}), we can provide additional input for the LLM to improve grading accuracy. There are 2 types of tests supported in pyevalai:
\begin{enumerate}
    \item \textbf{Text:} These tests consist of yes-no questions and corresponding points that should be granted if the LLM replies with yes or no (for example "Is the student cheating by using prohibited libraries in his code?" or "Did the student document the code properly?"). This allows administrators to specify certain evaluation criteria more clearly.
    \item \textbf{Code:} To assist LLMs in distinguishing between correct and incorrect code, code-tests consist of a question (e.g. "Does the function return correct values?") and a test function that takes the student's submitted code as input. After testing the student code, the test function returns a reply to the initial question (e.g. "The function returns correct values!") and a number of points that should be taken into account for the final score. For security reasons, the student code is not executed on the server but inside the student's notebook. To this end, the websocket connection is used to transmit input and return values between the student notebook and the tornado server.
\end{enumerate}

After the execution of these tests, the specified questions, corresponding replies, as well as the sum of achieved test-points are given as additional input to the language model to achieve better informed grades.

\section{Evaluation}

To validate PyEvalAI, 
we conducted a case study in a university level numerics course where volunteer Bachelor of Science in Computer Science students used interactive sample questions to prepare for the course exam. 
To obtain quantitative results about the AI's grading accuracy, we manually checked and rectified the AI generated feedback for all handed in solutions. 
Furthermore, we collected qualitative data through an anonymous feedback survey.
In the following section, we present our key findings.


\subsection{Quantitative}

In our case study, 20 volunteer students participated by solving exercises from a pool of 19 practice tasks. 
In total, they handed in 277 solutions, which were graded by both the AI and human tutors. These solutions resulted from varying numbers of attempts: 113 exercises were solved in a single attempt, 43 required two attempts, and 26 took three attempts to reach the correct solution.
In the following, we investigate the accuracy of PyEvalAI and how it helped students to improve their scores over multiple attempts.

\subsubsection{Grading Accuracy}
\label{sec:grading_accuracy}

First, we evaluate the accuracy of the AI model by comparing its generated scores with scores of human numerics tutors. 
Figure \hyperref[fig:grading_comparison]{8a} shows human gradings and AI gradings as coordinates in a 2D diagram where the number of occurrences is indicated by the point sizes. 
The majority of all points is located at the top right corner where AI and humans agree to give a large fraction of the total points.
As indicated by the small purple dots scattered across the top of the plot, in some cases the AI gave the highest possible score even though the submission was incomplete (e.g. if the solution is correct but the calculations were not presented in the submission). 
Figure \hyperref[fig:grading_differences]{8b} depicts a histogram of differences between AI and human gradings. 
It shows that 182 (or 65.7\%) of all 277 human gradings where identical to the AI. 
Furthermore, the distribution has a mean of $-0.14\%$ with a standard deviation of 20.73\% indicating that our model does not expose a significant bias.
Nonetheless, variations between automated and human gradings are still present. To some extent, however, this is also typical of human gradings. 
In our study, we have not yet evaluated human variations, as this would significantly increase the workload for all tutors.
However, this remains an interesting aspect for future research.

Lastly, we compare how often human corrections occured for the scores and the feedback text of the AI.
Table \ref{tab:human_corrections} presents how often the tutors agreed with the AI, only changed the grading, only corrected the feedback text or changed both. 
In 57.8\% of all submissions we have full agreement between the tutors and the AI. 
In 25.6\% of all cases the tutors changed the AI score as well as the feedback text. 
The remaining 16.6\% contain the cases in which the AI response is somewhat close to the human grading, such that only either the grading or the text had to be adjusted.
In these cases the AI provides a baseline for the feedback on which the tutors can build on.

The results support our approach of using LLMs as a semi-automatic tool to assist human tutors. 
In summary, although the AI model still produces occasional incorrect gradings, it aligns very well with human assessments on average. This supports our approach of using LLMs as a semi-automatic tool to assist human tutors, reducing their overall grading workload.

\begin{table}[t]
\centering
\caption{Human corrections on AI scores and feedback texts.}
\label{tab:human_corrections}
\setlength{\tabcolsep}{4pt}
\begin{tabular}{ccc}
& keep AI feedback & fix AI feedback \\
\hline
keep AI score & 160 (57.8\%) & 22 (7.9\%) \\
fix AI score & 24 (8.7\%) & 71 (25.6\%) \\
\end{tabular}
\end{table}


\subsubsection{Improved student performance through multiple graded attempts}
\label{sec:student_improvement}

\begin{figure}[t]
\centering
\begin{minipage}{.42\textwidth}
  \centering
  \includegraphics[width=1\linewidth]{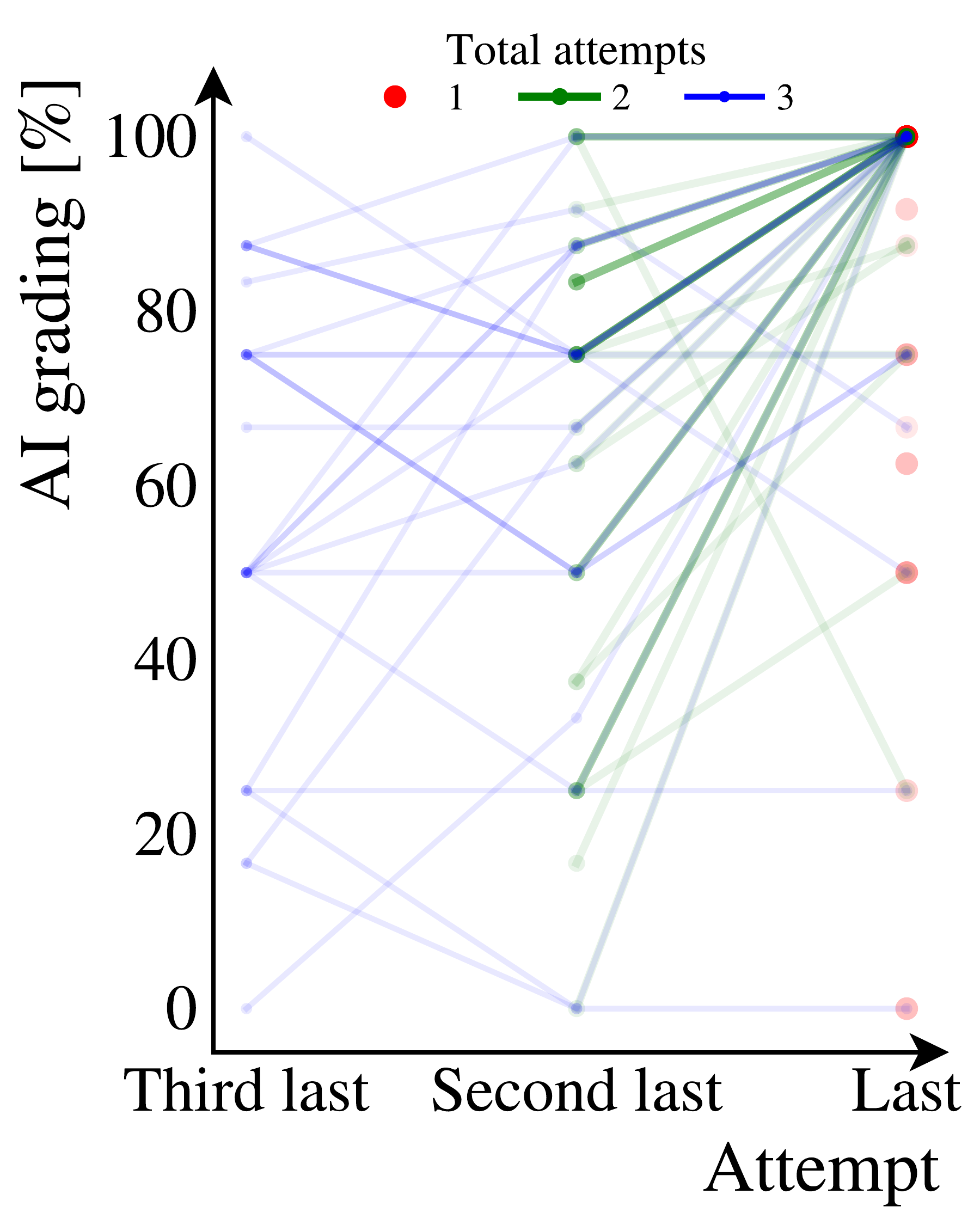}
  \captionof{figure}{Improvement of students on AI scores over multiple attempts.}
  \label{fig:student_improvement}
\end{minipage}%
\hspace{0.01\textwidth}
\begin{minipage}{.56\textwidth}
    \centering

    \captionof{table}{Average percentage of points per attempt.}
    \label{tab:percentage_per_attempt}
    \setlength{\tabcolsep}{4pt}
    \begin{tabular}{ccccc}
    Total & number of & $3^{rd}$ & $2^{nd}$ & \\
    attempts & solutions & last & last & Last \\
    \hline
    1 & 37 & & & 91.4\% \\
    2 & 20 & & 67.2\% & 95.4\% \\
    3 & 16 & 58.0\% & 63.3\% & 87.2\% \\
    \end{tabular}

    \captionof{table}{Number of students that performed better, equal and worse with multiple attempts.}
    \label{tab:n_improvements_per_attempt}
    \setlength{\tabcolsep}{4pt}
    \begin{tabular}{ccccc}
    \#attempts & from $\rightarrow$ to & better & equal & worse \\
    \hline
    2 & $1 \rightarrow 2$ & 36 & \phantom{0}6 & \phantom{0}1 \\
    3 & $1 \rightarrow 2$ & 11 & \phantom{0}5 & 10 \\
    3 & $2 \rightarrow 3$ & 19 & \phantom{0}5 & \phantom{0}2 \\
    \end{tabular}
\end{minipage}
\end{figure}

One important benefit of our system is the quick response by the LLM in comparison to traditional grading workflows which enables additional attempts to repeat and improve on tasks.
Figure \ref{fig:student_improvement} depicts the AI's grading of exercises over all attempts.
This time, the frequency of the data is visible as increasing opacity of the lines.
We again observe that a large number of last submissions are graded with 100\% of all points.
Moreover, when using multiple attemps, there is a significant improvement from the second last attempt to the last one.
To make this more clear, we compute the mean and standard deviation of all AI gradings for each attempt separately and report the numbers in Table \ref{tab:percentage_per_attempt}.
The last attempt always yields the most points with the lowest standard deviation. 
Students improve on average by 25\% -- 30\% compared to the second last attempt.
Table \ref{tab:n_improvements_per_attempt} displays the number of students that improved, worsened or stayed equal during their attempts. 
Also here we see that the vast majority of students indeed improved in grading.
Performing the same analysis on human grades instead of AI grades resulted in very similar statistics. 
Sometimes the students submitted solutions that were very similar to the previous attempts such that no change or even worsened gradings occured. 
Although our language model is configured to be deterministic, minor changes in the submission can still cause variations that may reduce the amount of points the model returns as the final grading.
This property is part of the LLM, which is difficult to prevent and requires further development in the future.


\subsection{Qualitative}

\begin{figure}[t]
    \includegraphics[width=1\linewidth]{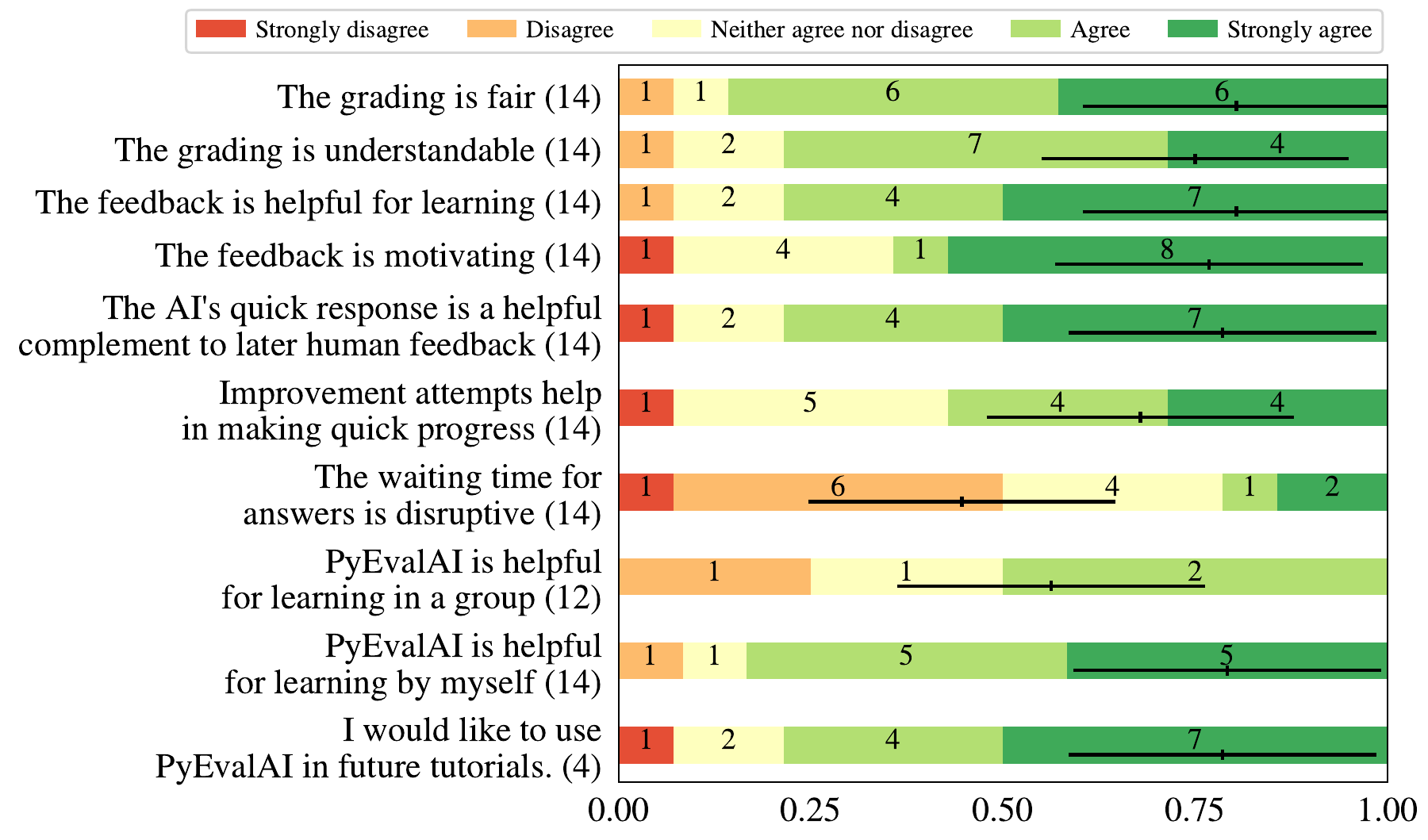}
    \caption{Results of the student survey regarding the experience with PyEvalAI.}
    \label{fig:bars_students_compressed}
\end{figure}

We conducted an anonymous survey in which 14 students participated.
The most important results are visualized in Figure \ref{fig:bars_students_compressed}.
In addition to the proportions of the responses, we compute a mean and standard deviation by converting all possible choices into equally-spaced numerical values between 0 (strongly disagree) and 1 (strongly agree) and visualizing these results as error bars for each question. 

We observe that PyEvalAI is well received by the students and that the grading is perceived as fair and understandable.
Similarly, the students acknowledge the quick AI feedback and consider it helpful, motivating and useful for making quick progress.
These results are in accordance with the quantitative evaluation in Section \ref{sec:grading_accuracy} for the gradings and the improvements with several attempts. 
Most students do not find the average waiting time of 88.2 seconds disruptive, but faster responses will be an obvious improvement to the PyEvalAI user experience. 
Finally, we ask how much benefit the AI tool provides when learning in groups versus when learning alone. 
Here, the students found PyEvalAI more useful when learning alone.
We suppose this could be due to group members already providing helpful feedback to each other, reducing the need for additional feedback from the AI.
In summary, most students rate PyEvalAI positively and would like to use it in future tutorials.

We also conducted a survey among the tutors with predominantly positive feedback. 
Although regular adjustments still had to be made by the tutors through the web-interface, 
all tutors agree or strongly agree that PyEvalAI is a helpful addition for students and confirm that the grading and feedback is mostly correct. 
This validates our findings during the quantitative evaluation in Section \ref{sec:grading_accuracy}.
However, the sample size (4 tutors) was too small to make meaningful statements about individual results. 
More data from further employment of PyEvalAI will increase statistical significance in future assessments. 

\section{Conclusion}

In this work, we presented PyEvalAI, a novel tool to provide students with immediate personalized AI generated feedback and support tutors in the grading process. PyEvalAI accepts text and code exercises, ensures privacy and is open source. A thorough evaluation of a case study in a numerics course demonstrates that the AI system often produces accurate and useful feedback that helps students to improve their solutions and tutors in the grading process.

In the future, PyEvalAI will be introduced in further courses to complement tutorials. 
A growing database of exercises, student solutions, AI and tutor feedback will enable thorough comparisons of different LLMs and help to improve prompting strategies and fine-tune models. 
We believe that, as local LLMs become increasingly more powerful and accurate, the learning experience for students can be further improved and tutors will be able to shift their focus from repetitive grading to providing individual feedback to students in the tutorials.

\bibliographystyle{splncs04}
\bibliography{mybibliography}
%

\end{document}